\documentclass[11pt,a4paper]{article}
\usepackage[hyperref]{emnlp2020}
\usepackage{times}
\usepackage{latexsym}

\usepackage{forest}
\usepackage{pbox}
\usepackage{adjustbox}
\usepackage{booktabs}
\usepackage{multirow}
\usepackage{array}
\usepackage{enumitem}
\usepackage{amsmath}
\usepackage{nameref} 
\usepackage{capt-of} 

\usepackage{microtype}

\aclfinalcopy 
\def\aclpaperid{2893} 

\def\newterm#1{\textit{#1}}
\def\paradigm#1{\textsc{#1}}

\newcommand{\dataset}{MSGS}

\newcommand*\subtxt[1]{_{\textnormal{#1}}}

\title{Learning Which Features Matter: RoBERTa Acquires a Preference for Linguistic Generalizations (Eventually)}

\author{\bf Alex Warstadt,$^{1}$ Yian Zhang,$^{2}$ Haau-Sing Li,$^{3}$ Haokun Liu,$^{3}$ Samuel R. Bowman$^{1,2,3}$ \\
$^{1}$Dept. of Linguistics, $^{2}$Dept. of Computer Science, $^{3}$Center for Data Science\\
New York University\\
Correspondence: \href{mailto:warstadt@nyu.edu}{\tt warstadt@nyu.edu}
}

\date{}

\begin{document}
\maketitle
\begin{abstract}

One reason pretraining on self-supervised linguistic tasks is effective is that it teaches models features that are helpful for language understanding. However, we want pretrained models to learn not only to represent linguistic features, but also to \emph{use} those features preferentially during fine-turning. With this goal in mind, we introduce a new English-language diagnostic set called MSGS (the Mixed Signals Generalization Set), which consists of 20 ambiguous binary classification tasks that we use to test whether a pretrained model prefers linguistic or surface generalizations during fine-tuning. We pretrain RoBERTa models from scratch on quantities of data ranging from 1M to 1B words and compare their performance on \dataset\ to the publicly available RoBERTa$\subtxt{BASE}$. We find that models can learn to represent linguistic features with little pretraining data, but require far more data to learn to \textit{prefer} linguistic generalizations over surface ones. Eventually, with about 30B words of pretraining data, RoBERTa$\subtxt{BASE}$ does demonstrate a linguistic bias with some regularity. We conclude that while self-supervised pretraining is an effective way to learn helpful inductive biases, there is likely room to improve the rate at which models learn which features matter.

\end{abstract}

\section{Introduction}

\begin{figure}[t]
    \centering
    \includegraphics[width=\columnwidth]{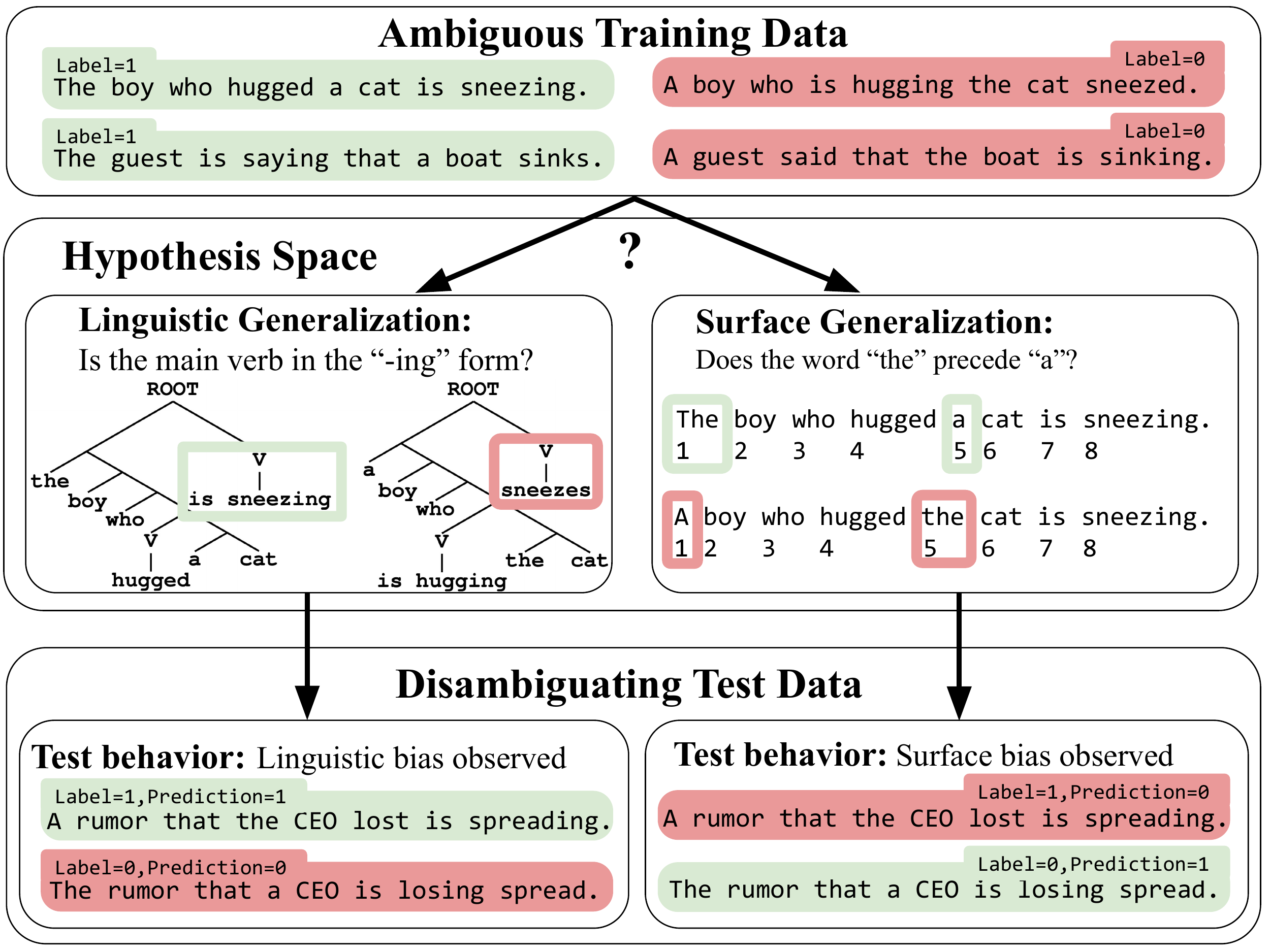}
    \caption{Example of an ambiguous experiment (without inoculation). A model is trained on ambiguous data whose labels are consistent with either a linguistic or a surface generalization, and tested on disambiguating data whose labels support only the linguistic generalization. Light green and dark red shading represents data or features associated with the positive and negative labels/predictions, respectively.}
    \label{fig:POS_example}
\end{figure}

Self-supervised pretraining through language modeling on massive datasets has revolutionized NLP. One reason this method works is that pretraining shapes a model's hypothesis space, giving it inductive biases that help it learn linguistic tasks \cite{howard2018universal}. Numerous probing studies have provided support for this idea by showing that language models learn representations that encode linguistic features \cite{gulordava2019colorless,tenney2019you,hewitt2019structural}. 

However, feature learning is just the first step to acquiring helpful inductive biases. Models must also be able to learn which features matter. The NLU datasets these models are often fine-tuned on are ambiguous and contain artifacts, and often support multiple possible generalizations. Neural networks are not mind readers: Models that have been shown to represent linguistic features sometimes fail to use them during fine-tuning on NLU tasks, instead adopting shallow surface generalizations \cite{jia2017adversarial,mccoy2019right}. To this end, recent work in probing pretrained models advocates for shifting the focus of study away from whether they represent linguistic features and in favor of whether they learn \emph{useful} representations of those features \cite{voita2020information,pimentel2020information,elazar2020amnesic}.

\begin{table*}[ht!]
    \centering
\resizebox{\textwidth}{!}{%
    \begin{tabular}{lllll}
    \toprule
        & \bf Feature type & \bf Feature description & \bf Positive example & \bf Negative example\\\midrule
        \multirow{5}{*}{\rotatebox[origin=c]{90}{\bf Surface}} 
        & Absolute position & Is the first token of S ``the''? & The cat chased a mouse. &  A cat chased a mouse. \\
        & Length & Is S longer than $n$ (e.g.,~3) words? & The cat chased a mouse. &  The cat meowed. \\
        & Lexical content & Does S contain ``the''? & That cat chased the mouse. &  That cat chased a mouse. \\
        & Relative position & Does ``the'' precede ``a''? & The cat chased a mouse. &  A cat chased the mouse. \\
        & Orthography & Does S appear in title case? & The Cat Chased a Mouse. & The cat chased a mouse. \\\midrule
        \multirow{4}{*}{\rotatebox[origin=c]{90}{\bf Linguistic}} 
        & Morphology & Does S have an irregular past verb? & The cats slept. &  The cats meow. \\
        & Syn. category & Does S have an adjective? & Lincoln was tall. & Lincoln was president. \\
        & Syn. construction & Is S the control construction? & Sue is eager to sleep. &  Sue is likely to sleep. \\
        & Syn. position & Is the main verb in ``ing" form? & Cats who eat mice are purring. & Cats who are eating mice purr. \\\bottomrule
    \end{tabular}%
}
    \caption{Schematic examples of the linguistic and  surface features in our experiments.}
    \label{tab:features}
\end{table*}

We investigate how RoBERTa \cite{liu2019roberta} acquires language-specific inductive biases during self-supervised pretraining. We track separately how RoBERTa's representation of linguistic features and its preferences for linguistic generalizations over surface generalizations change as the amount of pretraining data increases. We pretrain RoBERTa from scratch on datasets ranging from 1M to 1B words and evaluate these models alongside RoBERTa$\subtxt{BASE}$ in a series of experiments to probe the inductive biases of a pretrained model at the time of fine-tuning on a downstream task. 

We probe these models in three kinds of experiments: First, we conduct \emph{control} experiments where we fine-tune models on unambiguous binary classification tasks to test whether they learn to represent simple linguistic and surface features. Second, we conduct \emph{ambiguous} experiments following the \emph{poverty of the stimulus} design \cite{wilson2006learning}, as illustrated in Figure \ref{fig:POS_example}. In these experiments, we fine-tune a pretrained model on an ambiguous binary classification task in which the training set is consistent with both a linguistic generalization and a surface one. We then test the classifier on disambiguating data to reveal which generalization the model adopted, and by extension its preference among the two features. Third, we conduct \emph{inoculation} experiments \citep[following][]{liu2019inoculation} to test how hard it is to sway a model with a surface bias to adopt a linguistic generalization. We do this by introducing small amounts of disambiguating data into an otherwise ambiguous training set. We automatically generate data for all these tasks, and call the resulting dataset \dataset\ (Mixed Signals Generalization Set), pronounced ``messages''.

The results show that RoBERTa acquires a stronger linguistic bias as pretraining increases. RoBERTa$\subtxt{BASE}$ has the strongest linguistic bias, and requires little to no inoculating data to reliably make the linguistic generalization. In general, models with more pretraining data can generally be induced to adopt linguistic generalizations with less inoculating data. We also find a large gap between the amount of pretraining data that RoBERTa needs to learn the linguistic features necessary to generalize out-of-domain and the amount it needs to learns that it should \textit{prefer} those features when generalizing. The control experiments on unambiguous data reveal that models with little pretraining do actually represent the linguistic features, but nonetheless show a strong surface bias. In other words, the main contribution of pretraining to linguistic bias learning is devoted not to extracting features, but to learning which features matter. 

We conclude that helpful inductive biases can be learned through pretraining, but current models require abundant data to do so. The implications of this conclusion point in two directions: First, we can probably continue to pretrain on increasingly massive training sets to improve on the generalization and few-shot learning abilities of models like T5 \cite{raffel2019t5} and GPT-3 \cite{brown2020gpt3}. Second, since models learn useful features early, there is hope that future advances could accelerate by reducing the amount of data needed to learn which features matter. To aid in this effort, we release the MSGS dataset, our pretrained RoBERTas, and all our code: \href{https://github.com/nyu-mll/msgs}{\url{https://github.com/nyu-mll/msgs}}.

\section{Inductive Bias}

\paragraph{Background: Learning Inductive Bias} Any finite set of training examples shown to a learning algorithm like a neural network is consistent with infinitely many generalizable decision functions. Inductive biases are a learner's preferences among these functions. 
An inductive bias can eliminate certain possible functions altogether, or result in a preference for some over others \citep{haussler1988quantifying}.
For instance, an RNN classifier is capable of representing \textit{any} function, but prefers ones that focus mostly on local relationships within the input sequence \citep{dhingra2018neural,ravfogel2019studying}.

Some recent work seeks to design neural architectures that build in desirable inductive biases \cite{dyer2016recurrent,battaglia2018relational}, or compares the immutable biases of different architectures \cite{mccoy2020does,hu2020systematic}. However, inductive biases can also be \emph{learned} by biological \cite{harlow1949formation} and artificial systems alike \cite{lake2017building}. In the language model fine-tuning paradigm proposed by \citet{howard2018universal} and popularized by models such as BERT \citep{devlin2019bert}, a pretrained neural network plays the role of the learner. Pretraining adjusts a model's weights so that it will navigate the hypothesis space during training on a downstream task more effectively than a randomly initialized model.

There is a difference between learning to extract a linguistic feature and acquiring a bias towards using it when generalizing. There is ample evidence that BERT encodes features such as syntactic category and constituency \citep{tenney2019you,clark2019does,hewitt2019structural}. The acquisition of linguistic features is a \emph{prerequisite} for a linguistic bias. However, these findings do not tell us if the model will make use of these features to form generalizations during target task training, or if it will fall back on surface features that account for most of the data.

\begin{table*}[t]
    \centering
\resizebox{\textwidth}{!}{%
    \begin{tabular}{llrrl}
    \toprule
        \bf Dom. & \bf Split & $\textbf{L}_{\textbf{L}}$ & $\textbf{L}_{\textbf{S}}$ & \bf Sentence \\\midrule
\multirow{4}{*}{In} & \multirow{2}{*}{Train (Ambiguous)} & 1 & 1 & These men weren't hating that this person who sang tunes destroyed the vase.\\
  &  & 0 & 0 & These men hated that this person who sang tunes was destroying some vase.\\
  & \multirow{2}{*}{Inoc. (Disamb.)} & 1 & 0 & These men weren't hating that this person who sang tunes destroyed some vase.\\
  &  & 0 & 1 & These men hated that this person who sang tunes was destroying the vase.\\\midrule
\multirow{4}{*}{Out} & \multirow{2}{*}{Test (Disamb.)} & 1 & 0 & These reports that all students built that school were impressing some children.\\
  &  & 0 & 1 & These reports that all students were building the school had impressed some children.\\
  & \multirow{2}{*}{Aux. (Ambiguous)} & 1 & 1 & These reports that all students built the school were impressing some children.\\
  &  & 0 & 0 & These reports that all students were building that school had impressed some children.\\\bottomrule
         & 
    \end{tabular}%
}
    \caption{A full paradigm from the \paradigm{syntactic position $\times$ lexical content} task. $\textbf{L}_{\textbf{L}}$ and $\textbf{L}_{\textbf{S}}$ mark the presence of the linguistic feature (\emph{Is the main verb in the ``ing'' form?}) and surface feature (\emph{Does S contain ``the''?}), respectively. \emph{Dom.} is short for \emph{domain}.}
    \label{tab:paradigm}
\end{table*}

\paragraph{Methods: Measuring Inductive Bias}\label{sec:measuring}

We conduct three kinds of experiments to probe a model's preference for linguistic or surface generalizations: unambiguous control experiments, fully ambiguous experiments, and partially ambiguous inoculation experiments. Figure \ref{fig:POS_example} gives an overview of the ambiguous experiment design.

First, it only makes sense to compare a model's preference between two features if it actually represents both features. This is the goal behind \emph{control experiments}, in which we fine-tune RoBERTa to classify sentences based on a single linguistic or surface feature in a totally unambiguous setting.

Second, we conduct \emph{ambiguous experiments} on models that pass the controls. We fine-tune a pretrained model on a binary sentence classification task using \emph{ambiguous} data, which equally supports both a simple linguistic generalization and a simple surface one. For example, Figure \ref{fig:POS_example} shows a linguistic task where sentences in the positive class are defined by having a main verb in the ``ing'' form. We make the training data ambiguous by introducing a surface feature that distinguishes the two classes: In all (and only) training examples with label 1, the word ``the'' precedes the word ``a''. Based on this training data, a model could reasonably adopt either a linguistic generalization or a surface one.

We then test the classifier on disambiguating data to observe which generalization it made. In this kind of data, the labels align with the linguistic generalization, and contradict the surface one: For example, in Figure \ref{fig:POS_example}, ``a'' now always precedes ``the'' in the positive test examples with label 1. We quantify a model's inductive bias using a metric we call the \newterm{linguistic bias score} (LBS). We define LBS as the Matthews correlation between the model predictions and the labels on the disambiguating test set \cite{matthews1975correlation}.
If LBS is 1, the learner shows a systematic linguistic bias. If LBS is -1, it shows a systematic surface bias. If LBS is 0, it shows neither bias.

Finally, while the fully ambiguous experiments probe models' biases in an idealized setting, training data in more naturalistic contexts often does contain some evidence supporting a linguistic generalization over a simple surface one. To simulate this, we also conduct a series of \emph{inoculation experiments} \citep[following][]{liu2019inoculation}, in which we introduce small amounts of disambiguating data into an otherwise ambiguous training set. For each experiment, we replace 0.1\%, 0.3\%, or 1\% of the training data with examples that support the linguistic generalization and contradict the surface one. These experiments allow us to compare the strength of linguistic bias in models that show an overall surface bias: If two models adopt the surface generalization in the fully ambiguous case, we can still say that one has a stronger linguistic bias than the other if it requires less inoculation data to be swayed towards the linguistic generalization.

\section{Evaluation Data}

We introduce \dataset~(Mixed Signals Generalization Set), pronounced ``messages'', a dataset we design to be used in poverty of the stimulus and inoculation experiments. With the goal of contrasting inductive biases that are helpful and harmful in most NLP applications, the tasks in \dataset\ test a model's preferences for generalizations based on linguistic or surface features.

\paragraph{Features under Study}

Table \ref{tab:features} illustrates the 4 linguistic features and 5 surface features we consider.\footnote{We explored a slightly larger set of linguistic features and excluded several based on initial experiments showing our models did not encode them. For example, we constructed a task with the objective of identifying sentences that contain antonyms (e.g. \emph{The little girl likes the big dog.}), but found that only RoBERTa$\subtxt{BASE}$ could solve the unambiguous control task.} Each feature is meant to be representative of a broad category of features (e.g. morphological features), though the precise implementation of each feature is necessarily much narrower (e.g. \emph{Does the sentence have an irregular past verb?}). Forming generalizations based on surface features entails knowledge of the identity of certain words (in our case, only ``the'' and ``a''), the positional indices of words in the string, the total number of words in a string, or whether certain characters are lowercase or uppercase.\footnote{Although these are surface properties of the string, they are not all trivial for RoBERTa due to its subword tokenization.} Forming generalizations based on linguistic features requires more abstract knowledge of tense and inflectional morphemes, parts of speech, the control construction,\footnote{The \emph{control construction} is a syntactic construction in which a semantic argument of a predicate fills or \emph{controls} an argument slot of an embedded verb. The \emph{raising construction} is superficially similar, but the filler of the embedded argument slot is not a \emph{semantic} argument of the main predicate \cite{sag2003syntactic}. For instance, \emph{Sue is eager to sleep} is an example of control because the NP \emph{Sue} is the semantic subject of both \emph{eager} and \emph{sleep}. By contrast, \emph{Sue is likely to sleep} is an example of raising because \emph{Sue} is the semantic subject of \emph{sleep}, but not of \emph{likely}. These two phenomena have different syntactic derivations in some theories \cite{chomsky1981lectures}.} and hierarchical syntactic structures, none of which are encoded in the surface string.

\paragraph{Dataset Structure}\label{sec:dataset}

\dataset\ contains 20 ambiguous binary classification tasks each gotten by pairing one of 4 linguistic features with one of 5 surface features. We write $\textsc{feat}_1\times\textsc{feat}_2$ to denote a task that combines features $\textsc{feat}_1$ and $\textsc{feat}_2$. Each ambiguous dataset contains 50k sentences split into training, evaluation, and inoculation sets. \dataset\ also includes 9 unambiguous \newterm{control tasks}---one for each feature. Each control dataset contains 30k sentences split into training and evaluation sets.

For ambiguous tasks, we generate data in paradigms of 8 sentences following a $2\times 2\times 2$ design, as shown in Table \ref{tab:paradigm}. We vary the following three features: a binary linguistic feature, a binary surface feature, and the domain from which the sentence is sampled. We generate in-domain and out-of-domain sentences from different templates (see \S\ref{sec:data_generation}:\nameref{sec:data_generation} for more detail).

As shown in Table \ref{tab:paradigm}, we split the data into four contrasting pairs with different purposes: (1) \emph{Training data} is ambiguous in-domain data makes up 99\% to 100\% of the training set. (2) \emph{Inoculating data} is  disambiguating in-domain data which makes up 0.1\% to 1\% of the training set in experiments with inoculation. We show the classifier only the linguistic label ($\textbf{L}_{\textbf{L}}$) to nudge it towards adopting a linguistic generalization. (3) \emph{Test data} is disambiguating out-of-domain data used to test whether the model adopted the linguistic or surface generalization. (4) \emph{Auxiliary data} is ambiguous out-of-domain data used to test how well the model adapts to the out-of-domain templates, regardless of which generalization it makes.

For control tasks, we generate data in paradigms of 4 sentences following a $2\times 2$ design by varying the feature and domain. We use control tasks to test whether each pretrained model represents each feature well enough to fine-tune an effective classifier in an unambiguous setting.

\begin{figure*}[t]
    \centering
    \includegraphics[width=\textwidth]{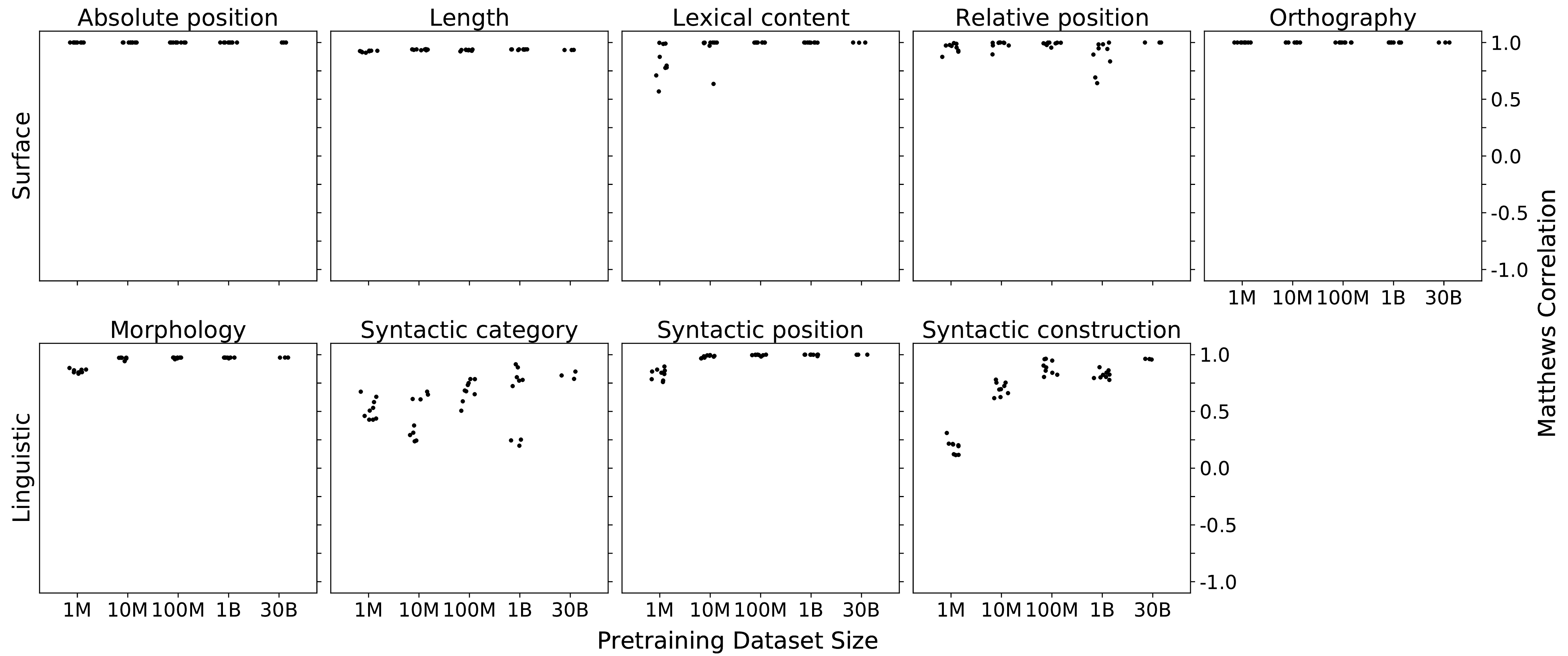} 
    \caption{Results on the main experiments measured in Matthews correlation for the surface control tasks (top) and linguistic control tasks (bottom). Note: For surface tasks a positive score represents a surface generalization.}
    \label{fig:control_results}
\end{figure*}

\paragraph{Data Generation}\label{sec:data_generation}

The data is generated from templates using a generation toolkit from \citet{warstadt2020blimp}. 
This toolkit includes a vocabulary of over 3000 entries labeled with grammatical features that allow for lexical variation in the data while maintaining grammatical well-formedness. Although generated sentences often describe unlikely or implausible scenarios (e.g., \emph{The lawyer was sinking all canoes}), semantic plausibility is independent of all the features we examine, so this should not affect a model that genuinely encodes these features. To prevent out-of-vocabulary tokens affecting our results, we ensure that every word stem in the vocabulary appears in the pretraining datasets for our RoBERTa models (see \S \ref{sec:pretraining_data}).

Our experimental logic only makes sense if we are reasonably confident that models can only achieve high test performance by genuinely adopting a linguistic generalization. However, training models on generated data can easily lead to overfitting, and classifiers trained and tested on data from the same domain can achieve perfect performance even on arbitrary tasks with random labels \cite{hewitt2019control}. For this reason, our primary evaluations test models' ability to \emph{generalize} out-of-domain. We manipulate domain in two ways:

First, we generate training data and test data for each dataset from separate in-domain and out-of-domain templates. 
Thus a model cannot succeed at test time simply by recognizing a template or a key part of a template. 
For example, in the \paradigm{syntactic position $\times$ lexical content} paradigm shown in Table \ref{tab:paradigm}, the in-domain data contrasts the main verb with a verb in a relative clause embedded in the complement clause of a verb; while the out-of-domain data contrasts the main verb with a verb in the complement clause of a noun. In most tasks, each domain itself is generated from multiple templates as well to widen the domain and encourage better generalization during training. 

Second, on tasks that test lexical knowledge (for instance, the knowledge that \emph{slept} is an irregular past verb and \emph{meow} is not), we divide the crucial lexical items into in-domain and out-of-domain sets. Thus, a model cannot succeed by memorizing the keywords associated with each class. See the Appendix for a more detailed description of the implementation details for each feature.

\begin{figure*}[t]
    \centering
    \includegraphics[width=\textwidth]{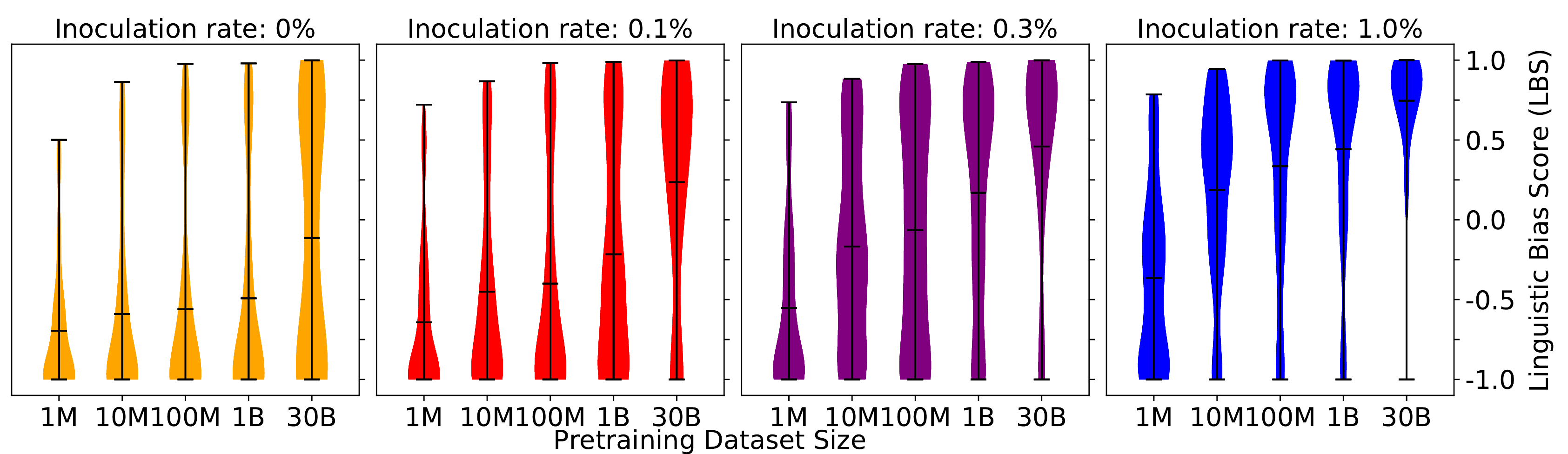}
    \caption{Results measured in LBS for each pretraining and inoculating data amount, aggregated over the 20 tasks in \dataset. We exclude models that fail the corresponding controls, as described in Section \ref{sec:results}. High density near LBS of 1 means many models in that group have a linguistic bias; high density near -1 means many models have a surface bias. Models with stronger linguistic bias achieve higher LBS with less inoculation data.}
    \label{fig:violin_plot}
\end{figure*}

\section{Models, Pretraining, \& Fine-Tuning} 
    
We test 13 RoBERTa models in our main experiments: We pretrain 12 from scratch, and also test RoBERTa$\subtxt{BASE}$ pretrained by \citet{liu2019roberta}.

\subsection{Pretraining}\label{sec:pretraining_data}

\paragraph{Pretraining Data}
We pretrain RoBERTa using scaled-down recreations of the dataset used by \citet{devlin2019bert} to train BERT, i.e
English Wikipedia (2.5 billion tokens) and BookCorpus (800 million tokens). Both are included in the RoBERTa pretraining data.\footnote{RoBERTa uses English Wikipedia, BookCorpus, CC-News, OpenWebText, and STORIES in pretraining.}
We download the latest Wikipedia dump as of Feb 1, 2020. The original BookCorpus \citep{zhu2015aligning} is no longer available, so we collect similar data from Smashwords, the original source of BookCorpus.\footnote{We collect our data using the Wikipedia XML dump \href{https://dumps.wikimedia.org/mirrors.html}{https://dumps.wikimedia.org/mirrors.html} and data-processing code \href{https://github.com/attardi/wikiextractor}{https://github.com/attardi/wikiextractor}, and a Smashwords crawler \href{https://github.com/soskek/bookcorpus}{https://github.com/soskek/bookcorpus}.}

We pretrain RoBERTa on four training sets containing different numbers of  words: 1M, 10M, 100M, and 1B.\footnote{The publicly available RoBERTa$\subtxt{BASE}$ is trained on 160GB of data, which we estimate to be about 30B words.} To make these datasets, we sample entire Wikipedia articles and Smashwords books independently, keeping the proportions of Wikipedia and Smashwords text approximately constant.

\paragraph{Model Sizes}
Model size is the only hyperparameter we systematically search over during pretraining. We consider smaller model sizes to prevent overfitting on small training sets. The detailed configurations of the model sizes are summarized in the Appendix. We use RoBERTa$\subtxt{BASE}$ from \citet{liu2019roberta} as our largest model size. The other configurations represent a scale roughly based on settings used in  \citet{sanh2019distilbert}, \citet{vaswani2017attention}, \citet{jiao2019tinybert}, and \citet{tsai2019small}.

\paragraph{Search Range} For dropout, attention dropout, learning rate decay, weight decay and the Adam parameters, we adopt the same parameter values used in \citet{liu2019roberta}. We fix warm up steps to be 6\% of max steps, peak learning rate to be 5e-4, early stopping patience to be 100M tokens, and heuristically define the search range of model size, max steps and batch size for each training set.

\paragraph{Search Results} 
We randomly sample hyperparameters from the search range and train 25 models for each of the 1M, 10M, 100M datasets. We train 10 models on the largest (1B) dataset due to resource limitations. For each training set size, we choose three of the resulting models to evaluate. In order to avoid confounds caused by different model sizes, for each training set we choose three models of the same size that have the lowest perplexity. The hyperparameters and validation perplexities of the selected models are listed in the Appendix.

\subsection{Fine-Tuning}

We loosely follow the hyperparameter settings that \citet{liu2019roberta} used for fine-tuning on GLUE tasks \cite{wang2018glue}, and use the following learning rates: \{1E-5, 2E-5, 3E-5\}. We depart from \citeauthor{liu2019roberta}\ in using a batch size of 16 and training for 5 epochs without early-stopping in all runs. These changes are based on pilots that showed that larger batch sizes and longer fine-tuning were no more effective for our tasks.

We conduct 3,471 fine-tuning runs: We fine-tune 13 RoBERTa models: (3 random initializations) $\times$ (4 pretraining data amounts) $+$ (1 RoBERTa$\subtxt{base}$). 
We fine-tune each model 267 times: (3 learning rates) $\times$ ((9 control tasks) $+$ (20 ambiguous tasks) $\times$ (4 inoculation amounts)). 
We evaluate model performance using LBS (see \S \ref{sec:measuring}:\nameref{sec:measuring}).

\begin{figure*}[t]
    \centering
    \includegraphics[width=\textwidth]{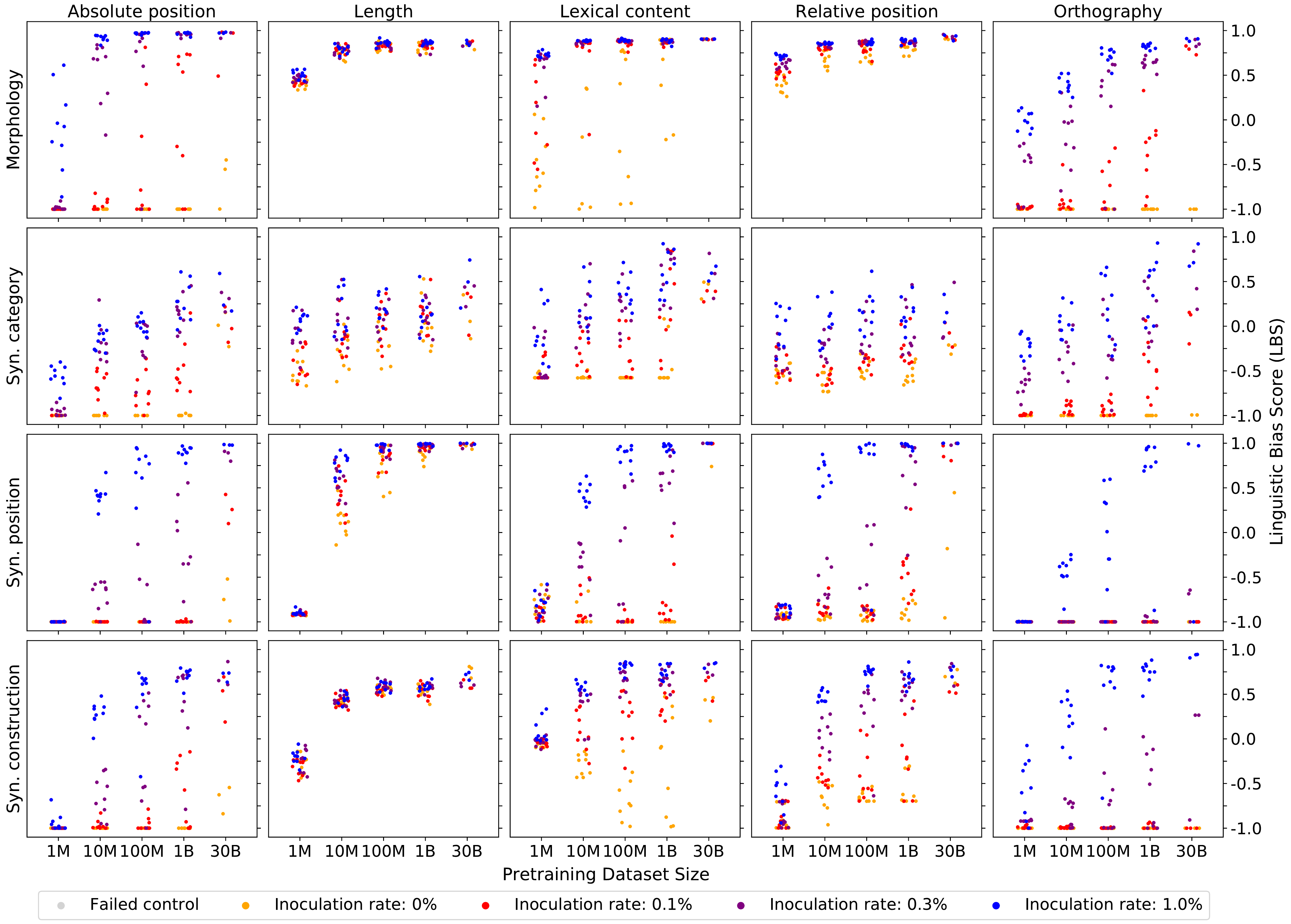}
    \caption{Results of the ambiguous binary classification tasks measured in LBS for every (linguistic feature, surface feature) pair. Each plot in the matrix shows the results on the disambiguating test items after training on an ambiguous task. All experiments on the same row investigate the same linguistic feature; all experiments on the same column investigate the same surface feature. Each data point represents one run. The x-axis of the point is the pretraining size of the model, and the y-axis is its LBS. Models with stronger linguistic bias achieve higher LBS with less inoculation data. Gray points show runs where the corresponding controls did not pass. A black-and-white version of this figure separating color channels into separate plots, can be found in the Appendix.}
    \label{fig:main_results}
\end{figure*}

\section{Results \& Discussion}\label{sec:results}

We have several main findings: (1) models learn to \emph{represent} both surface features and linguistic features with relatively little data; (2) RoBERTa begins to acquire a linguistic bias with over 1B words of pretraining data; (3) increasing pretraining data strengthens linguistic bias; (4) there is considerable variation in models' preferences between specific pairs of linguistic and surface features.

\paragraph{Control results} Figure \ref{fig:control_results} shows the results for the controls. Performance is near ceiling for most models and features. Because we evaluate all the models out-of-domain, this result cannot be explained by the models simply memorizing the features from the task training data. Thus, we conclude that most pretrained models we test encode both linguistic and surface features. 

The only exceptions are the syntactic category and syntactic construction features, for which models with less than 100M perform poorly. In subsequent plots, we filter out results where the controls are not passed. Specifically, if a particular combination of model checkpoint and learning rate achieves a Matthews correlation of less than 0.7 on the control task for feature $F$, we eliminate all results with this combination for any task involving $F$ in Figure \ref{fig:violin_plot}, or represent them as gray points in Figure \ref{fig:main_results}.

\paragraph{Main Experiment Results}

Figure \ref{fig:violin_plot} summarizes the main experiment results. For a given amount of pretraining and inoculation data, we consider all classifiers trained on all 20 tasks in MSGS and plot the density of their linguistic bias scores (LBSs).

The results in the leftmost box (with 0\% inoculation) show that only RoBERTa$\subtxt{BASE}$ demonstrates a consistent linguistic bias in the fully unambiguous setting. That said, it still adopts the surface bias much of the time. The other models show a clear surface bias overall. The results of experiments with inoculation data show that models with more pretraining data require less inoculation data to be swayed towards the linguistic generalization. We consistently observe, for each pretraining quantity, a phase transition where the linguistic generalization begins to overtake the surface generalization upon exposure to a certain amount of inoculating data. For example, the 1B model goes through this transition between 0.1\% and 0.3\% inoculating data. The 100M and 10M models go through this transition between 0.3\% and 1\% inoculating data. The phase transition comes earlier for models with more pretraining, indicating they have a stronger linguistic bias. We also notice distinctive behavior for the models at the extreme ends of pretraining data quantity: The 1M model never completes the transition, suggesting it has a strong surface bias, and RoBERTa$\subtxt{BASE}$ appears to be in the middle of this transition with 0\% inoculating data, suggesting that even more pretraining data could produce a model with a more consistent linguistic bias.

These findings are echoed in individual task results in Figure \ref{fig:main_results}.\footnote{Analogous results for the held out training-condition data, inoculation data, and auxiliary data are in the Appendix.} In each plot, models with the same amount of inoculation data (i.e. points with a given color) have higher LBS as the amount of pretraining data increases. Notably, on ambiguous tasks involving \paradigm{lexical content}, RoBERTa$\subtxt{BASE}$ usually favors generalizations based on linguistic features without any inoculating data, which no other pretrained model does. We find this result quite striking: Even if the labels are perfectly correlated with the presence or absence of the word ``the'', RoBERTa$\subtxt{BASE}$ overlooks that fact in favor of a deeper generalization based on an abstract feature like the inflectional form of a verb in a particular syntactic position. Furthermore, this preference is clearly \emph{acquired} through additional pretraining. The results for \paradigm{morphology $\times$ orthography} is a typical illustration of the differences between models. The 1M model never adopts the linguistic generalization based on the morphological feature, though it eventually rejects the surface generalization. The 100M and 1B models make robust linguistic generalizations only with 1.0\% inoculating data. By contrast, RoBERTa$\subtxt{BASE}$ requires only 0.1\% inoculating data (i.e.~10 out of 10k examples) to adopt the linguistic generalization. 

\paragraph{Surface Biases of RoBERTa} Our results also suggest some specific conclusions about which kinds of surface features RoBERTa pays attention to.\footnote{\dataset\ does not come close to representing the full range of possible relevant lexical or syntactic features, preventing us from making strong conclusions about which specific linguistic features RoBERTa has biases in favor of.} For instance, these models have little preference for sentence length. As shown in the second column of Figure \ref{fig:main_results}, most of the models form linguistic generalizations rather than generalizations based on sentence length, even with no inoculating data. By contrast, the models strongly prefer generalizations based on orthography---and to a lesser extent lexical content and word order---over linguistic generalizations.

\paragraph{The Success of Pretrained Models} Our findings provide insight into why pretraining on massive datasets is so successful. While linguistic feature learning is a major effect of pretraining, it is far from the end of the story: Pretraining also helps models learn which features are central to language. However, this second kind of learning seems to require far more exposure to data with current models and pretraining techniques. Therefore, massive datasets are needed to teach models which features are useful for generalizing.

The data scale at which we observe RoBERTa beginning to show a linguistic bias (between 1B and 30B words) is similar to the amount of pretraining data used by the first pretrained LMs to achieve major successes at NLU tasks, such as ELMo \cite{peters2018elmo} and BERT \cite{devlin2019bert}. This suggests a crucial data threshold below which language model pretraining is unlikely to be significantly helpful for most applications with current model architectures, and may explain the many-year gap between the development of neural LMs and the first major applications of LM pretraining: The early LMs must have not have been trained sufficiently to cross that threshold, yielding consistently poor results.

\section{Related work}

There is increasing interest in studying the inductive biases of neural networks. Much of this work has grown out of numerous findings that these models often fail to generalize in ways that task designers intend. For example, \citet{jia2017adversarial} and \citet{mccoy2019right} demonstrate that ambiguity in widely used NLU datasets like SQuAD \cite{rajpurkar2016squad} and MultiNLI \cite{williams2018broad} leads models like BERT to adopt some surface generalizations, despite the fact that they represent linguistic features. This continues to be a problem for models like RoBERTa$_{\subtxt{BASE}}$ which show an overall linguistic bias in our experiments. However, for tasks like NLI, the underlying linguistic feature depends on a combination of significant syntactic knowledge, semantic knowledge, and world knowledge. It stands to reason that representations and preferences for such high level features require more data to learn than the features we probe.

Other work has used the poverty of stimulus design to study inductive biases associated with particular neural architectures during syntactic generalization. \citet{ravfogel2019studying} train RNNs on a morphological prediction task using artificial languages derived from naturally occurring English text, finding that RNNs show a recency bias in acquiring agreement rules. \citet{mccoy2018revisiting,mccoy2020does} train a seq2seq models on generated data ambiguous between a surface and a structural generalization to learn the subject-auxiliary inversion rule in English question formation. They find that, while tree-structured models show a structural bias, sequence models do not. \citet{warstadt2020can} conduct related experiments on subject-auxiliary inversion and other English structural rules, and find that BERT likely acquires a structural bias from pretraining. 

More abstract inductive biases have also been studied. Using zero-shot learning in an artificial language, \citet{lake2018generalization} show that RNNs lack a bias in favor of learning compositional meanings for new symbols. \citet{gandhi2019mutual} and \citet{gulordava2020one} explore conditions under which neural networks exhibit a bias towards learning mutually exclusive meanings for new symbols.

Data augmentation and inoculation have also been explored previously as a way to influence how models generalize. \citet{mccoy2019right} and \citet{min2020syntactic} show that small amounts of inoculating data during training on textual entailment help BERT overlook certain surface generalizations. \citet{jha2020does} study inoculation using a constructed language of numerical sequences. Like us, they generate ambiguous datasets, though they only compare features that resemble our surface features. They find that it is relatively easy to nudge models away from shallow generalizations, but harder to nudge them towards deeper ones.

Finally, several earlier studies explored how increasing training data impacts linguistic knowledge in LMs. Unlike the present study, these studies evaluate LMs using an unsupervised acceptability judgment task on minimal pairs (i.e.~not during fine-tuning), and do not attempt to separate feature learning from feature preferences. \citet{vanschijndel2019quantity} find the greatest increase in sensitivity to acceptability contrasts occurs between training on 2M and 10M words. \citet{warstadt2020blimp} find that while LMs learn agreement phenomena at a similarly early stage, other phenomena require more data to learn. Finally, \citet{hu2020systematic} find that adopting architectures that build in linguistic bias, such as RNNGs \cite{dyer2016recurrent}, has a bigger effect on the acceptability task than increasing training data from 1M to 40M words.

\section{Future Work \& Conclusion}

Our experiments shed light on the relationship between pretraining data and an inductive bias towards linguistic generalization. Our results indicate that, although some abstract linguistic features are learnable from relatively small amounts of pretraining data, models require significant pretraining after discovering these features to develop a bias towards \emph{using} them when generalizing. This gives some insight into why extensive pretraining helps general-purpose neural networks adapt to downstream tasks with relative ease. 

We also introduce \dataset, a new diagnostic dataset for probing the inductive biases of learning algorithms using the poverty of the stimulus design and inoculation, and also introduce a set of 12 RoBERTa models we pretrain on smaller data quantities. These models could prove to be a helpful resource for future studies looking to study learning curves of various kinds with respect to the quantity of pretraining data.

Finally, while our results naturally lead to the conclusion that we should continue to pursue models with ever more pretraining, such as GPT-3 \cite{brown2020gpt3}, we do not wish to suggest that this will be the only or best way to build models with stronger inductive biases. Future work might use \dataset\ as a diagnostic tool to measure how effectively new model architectures and self-supervised pretraining tasks can more efficiently equip neural networks with better inductive biases.

\section*{Acknowledgments}

This project has benefited from financial support to SB by Eric and Wendy Schmidt (made by recommendation of the Schmidt Futures program), by Samsung Research (under the project \textit{Improving Deep Learning using Latent Structure}), by Intuit, Inc., and in-kind support by the NYU High-Performance Computing Center and by NVIDIA Corporation (with the donation of a Titan V GPU). This material is based upon work supported by the National Science Foundation under Grant No. 1850208 and 1922658. Any opinions, findings, and conclusions or recommendations expressed in this material are those of the author(s) and do not necessarily reflect the views of the National Science Foundation. 

\bibliography{emnlp2020}
\bibliographystyle{acl_natbib}

\clearpage
\appendix

\section{Data Description}

\begin{table*}[ht]
    \centering
\resizebox{\textwidth}{!}{%
    \begin{tabular}{lllll}
    \toprule
        & \bf Feature type & \bf Feature description & \bf Positive example & \bf Negative example\\\midrule
        \multirow{5}{*}{\rotatebox[origin=c]{90}{\bf Surface}} 
        & Absolute position & Is the first token of S ``the''? & The cat chased a mouse. &  A cat chased a mouse. \\
        & Length & Is S longer than $n$ (e.g.,~3) words? & The cat chased a mouse. &  The cat meowed. \\
        & Lexical content & Does S contain ``the''? & That cat chased the mouse. &  That cat chased a mouse. \\
        & Relative position & Does ``the'' precede ``a''? & The cat chased a mouse. &  A cat chased the mouse. \\
        & Orthography & Does S appear in title case? & The Cat Chased a Mouse. & The cat chased a mouse. \\\midrule
        \multirow{4}{*}{\rotatebox[origin=c]{90}{\bf Linguistic}} 
        & Morphology & Does S have an irregular verb? & The cat slept. &  The cat meows. \\
        & Syn. category & Does S have an adjective? & Lincoln was tall. & Lincoln was president. \\
        & Syn. construction & Is S the control construction? & Sue is eager to sleep. &  Sue is likely to sleep. \\
        & Syn. position & Is the main verb in ``ing" form? & Cats who eat mice are purring. & Cats who are eating mice purr. \\\bottomrule
    \end{tabular}%
}
    \caption{Schematic examples of the linguistic and surface features.}
    \label{tab:features}
\end{table*}

\dataset\ contains 5 surface features and 4 linguistic features, summarized in Table \ref{tab:features} (repeated from the main body of the paper, for convenience). Implementation details for the features are described below. The implementation of one feature sometimes depends on other feature it is paired with in an ambiguous dataset.

\paragraph{Absolute position} This feature is 1 \emph{iff} the sentence begins with the word ``the''. We generally ensure that sentences bearing a value for this feature contains two clauses and four determiners total. Some sentences in \paradigm{syntactic category $\times$ absolute position} contain fewer than four NPs. The in-domain and out-of-domain sentences differ in the order or position of the clauses.

\paragraph{Length} This feature is 1 \emph{iff} the sentence exceeds some number of tokens. The exact threshold varies depending on the linguistic feature in an ambiguous task, since different linguistic features lead to sentences of different length, on average. In mixed tasks, we vary the length of sentences by adjoining subordinate clauses (e.g. \emph{If Sue wakes}) of varying length to the clause in which the linguistic feature is varied.

\paragraph{Lexical content} 
This feature is 1 \emph{iff} the sentence contains \emph{the}. 
The sentences generally contain at least two clauses and four determiners. The position of \emph{the} varies between in-domain and out-of-domain sentences.

\paragraph{Relative position} 
This feature is 1 when \emph{the} precedes \emph{a}, and 0 when \emph{a} precedes \emph{the}.
The sentences generally contain at least two clauses and four determiners. Thus, there are six different configurations in which \emph{the} precedes \emph{a}, and these are separated into in-domain and out-of-domain templates.

\paragraph{Orthography} 
This feature is 1 \emph{iff} the sentence appears in title case. 
In the control paradigm, the sentences generally contain two clauses, whose positions are varied between in-domain and out-of-domain sentences. 

\paragraph{Lexical semantics}
This feature is 1 \emph{iff} the sentence contains a pair of antonyms.
In sentences with label 0, there is a pair of words in a hypernym/hyponym or synonym relation. There are 21 pairs of adjective antonyms and 21 pairs of verb antonyms (not accounting for different inflectional forms). To prevent the task being solvable using lexical content, these pairs are divided into in-domain and out-of-domain sets. There are different templates corresponding to whether the antonyms are adjectives, intransitive verbs, or transitive verbs. Each template appears in both in-domain and out-of-domain sentences.

\paragraph{Morphology} 
This feature is 1 when the sentence contains an irregular past tense verb, and 0 when it contains a 3rd person present plural verb (identical to the bare form).
Sentences either contain an irregular past tense verb or a regular 3rd person present plural verb (identical to the bare form). We do this because other verb forms can be identified by inflectional morphemes such as \emph{-s} or auxiliaries such as \emph{have}, and so discrimination between them could in some cases reduce to a lexical content task. The verbs are divided into in-domain and out-of-domain sets.

\paragraph{Syntactic category} 
This feature is 1 \emph{iff} the sentence contains an adjective. 
To diversify the templates, we consider all grammatical combinations of a noun, an adjective, a locative PP, and a proper name (e.g.,~\emph{Sue is the tall actress in the park}, or \emph{The actress is Sue}). In out-of-domain sentences we also include single-word nominal predicates like \emph{president} (see the example in Table \ref{tab:features} to control for the fact that predicative adjectives are always a single, lowercase word. This gives a total of 19 templates divided into in-domain and out-of-domain sets, some with adjectives and some without. The set of adjectives is also split between domains.

\paragraph{Syntactic construction}
This feature has value 1 \emph{iff} the sentence contains the control construction. In the control construction a semantic argument of a predicate fills or \emph{controls} an argument slot of an embedded verb \cite{sag2003syntactic}. For instance, in \emph{Sue is eager to sleep}, the NP \emph{Sue} surfaces as the syntactic subject of \emph{eager}, but \emph{Sue} is also understood as the semantic subject of \emph{sleep}. This contrasts with the \emph{raising} construction in \emph{Sue is likely to sleep}, where \emph{Sue} is again surfaces as the syntactic subject of \emph{likely} in the main clause, and is the semantic subject of \emph{sleep} in the embedded position, but is not a semantic argument of \emph{likely}. Different predicates are compatible with control and raising: \emph{eager} is a control predicate and \emph{likely} is a raising predicate. We include control and raising predicates of three kinds: subject control/raising verbs, object control/raising verbs, and control/raising adjectives. Specific predicates are divided into in-domain and out-of-domain sets, but all three kinds of predicates appear in both domains.

\paragraph{Syntactic position}
All sentences contain at one or two embedded clauses. We include six sentence types, divided into in-domain and out-of-domain. For example, some sentences contain a relative clause within a relative clause, or a verb phrase with a complement clause. Each sentence type is generated from multiple templates varying the position of the clauses. The set of \emph{-ing} verbs is not split between domains.

\clearpage

\section{Pretraining Details}

\begin{minipage}[t]{1.0\textwidth}
    \centering
    \begin{tabular}{lrrrrr}
    \toprule
    \bf Name & \bf L & \bf AH& \bf HS& \bf FFN &\bf P\\
    \midrule
    Base & 12 & 12 & 768 & 3072 & 125M\\
    Med & 6 & 12 & 768 & 3072 & 82M\\
    Med-Small & 6 & 8 & 512 & 2048 & 45M\\
    Small & 4 & 8 & 384 & 1200 & 26M\\
    XSmall & 3 & 4 & 256 & 1024 & 15M\\
    \bottomrule
    \end{tabular}%
    \captionof{table}{The RoBERTa model sizes we search over during pretraining. AH = number of attention heads; HS = hidden size; FFN = feed-forward network dimension; P = number of parameters.}
    
    \vspace{0.5in}
    
    \begin{tabular}{lrrrr}
    \toprule
    \bf Training Size& \bf Model Size & \bf Max Steps & \bf Batch Size & \bf Validation Perplexity \\
       \midrule
    1B & \sc{base} & 31K & 4096 & 3.84 \\
    1B & \sc{base} & 100K & 512 & 3.93 \\
    1B & \sc{base} & 31K & 1024 & 4.25 \\
    \midrule
    100M & \sc{base} & 31K & 1024 & 4.61 \\
    100M & \sc{base} & 100K & 512 & 4.99 \\
    100M & \sc{base} & 31K & 512 & 5.02 \\
    \midrule
    10M & \sc{base} & 10K & 512 & 10.78\\
    10M & \sc{base} & 10K & 1024 & 11.31\\
    10M & \sc{base} & 31K & 512 & 11.58\\
    \midrule
    1M & \sc{med-small} & 10K & 512 & 134.18\\
    1M & \sc{med-small} & 31K & 512 & 139.39\\
    1M & \sc{med-small} & 100K & 512 & 153.38\\\bottomrule
    \end{tabular}
    \captionof{table}{The pretraining parameters of the 12 models we use in our experiments. }
\end{minipage}
\newpage

\section{Additional Results}

\begin{minipage}[t]{1.0\textwidth}
    \centering
    
    \includegraphics[width=\textwidth]{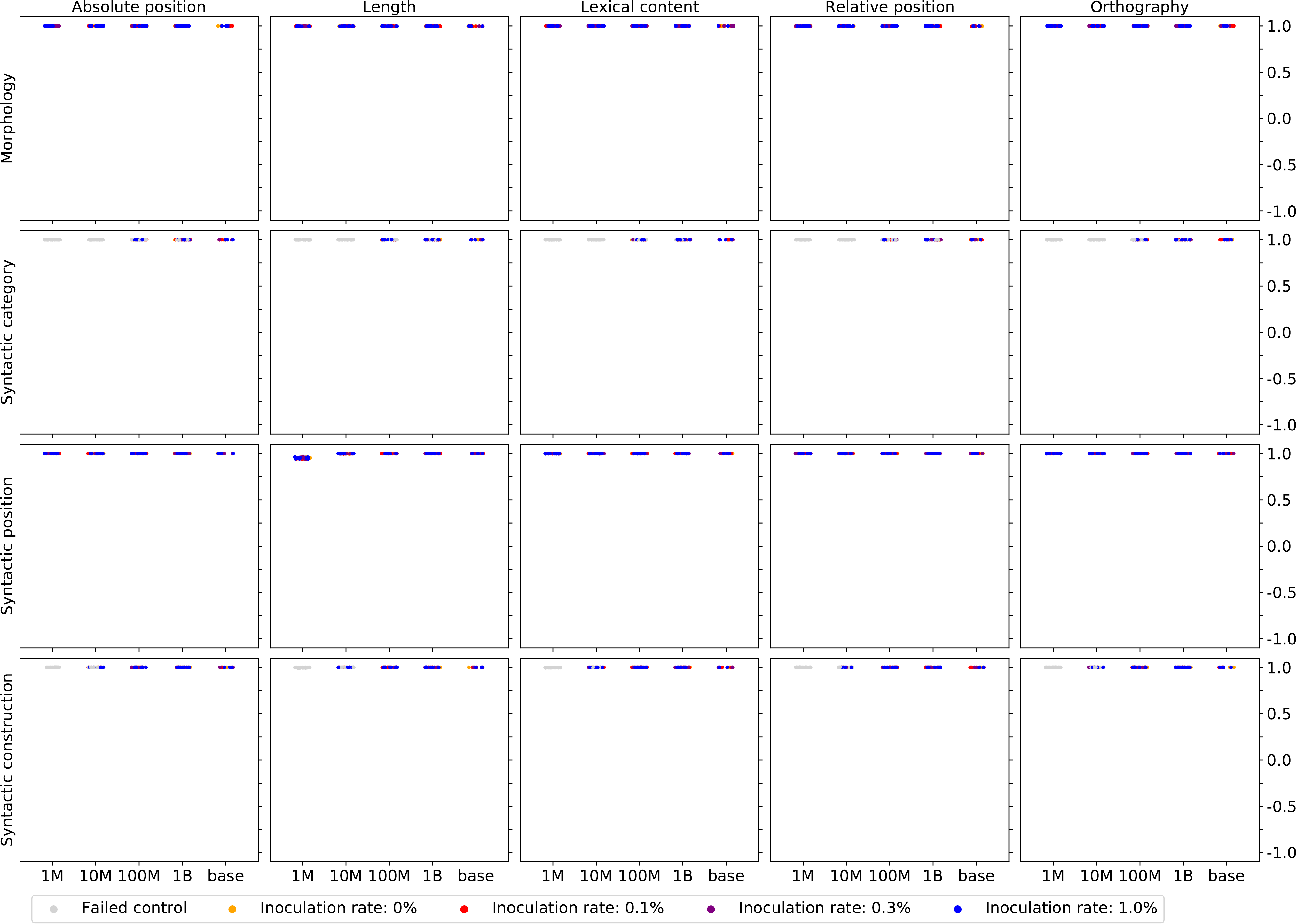}
    \captionof{figure}{Results on the held-out training-condition items (in-domain/mixed) measured in LBS.}
    \label{fig:train_results}
    \centering
    \includegraphics[width=\textwidth]{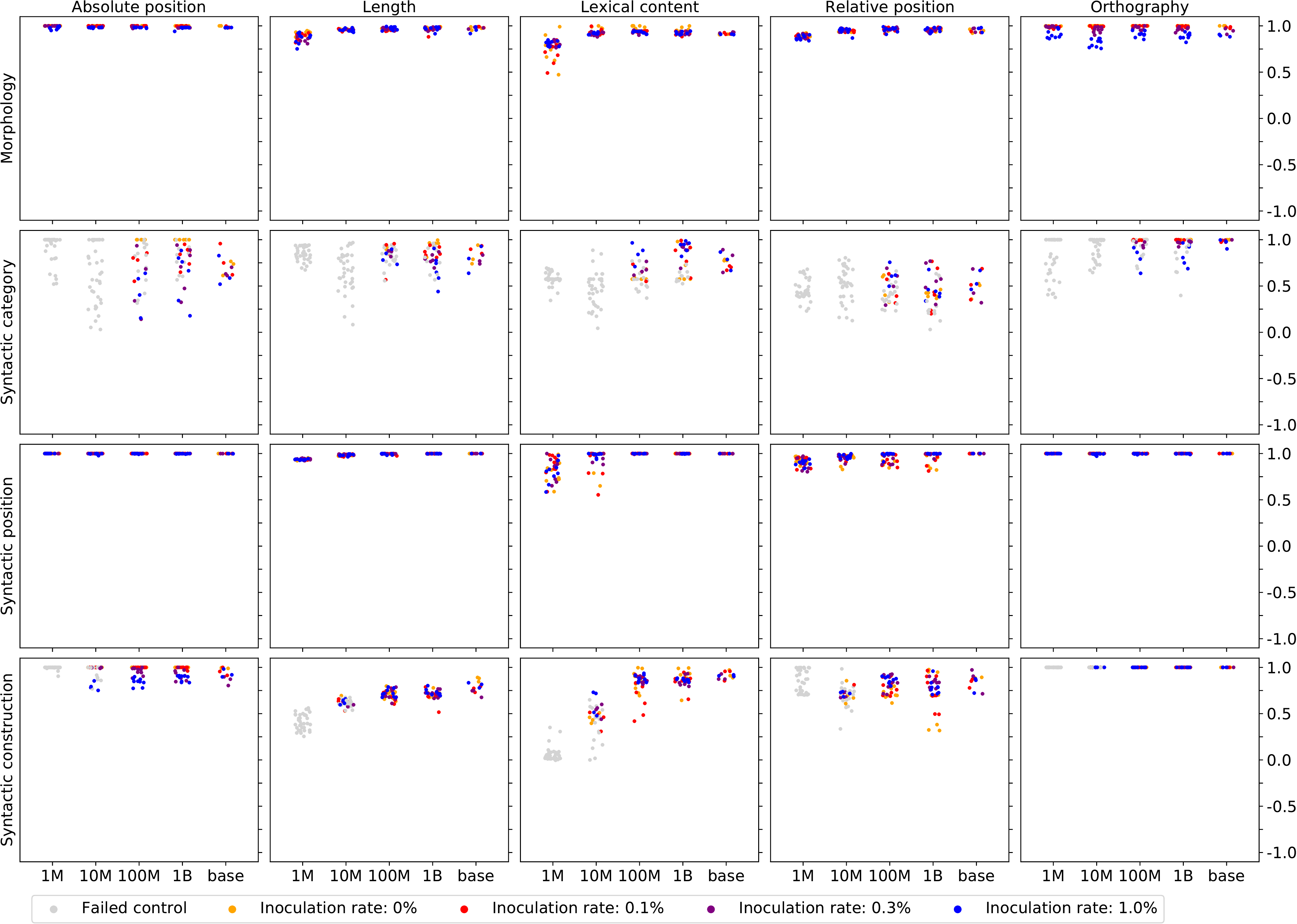}
    \captionof{figure}{Results on the held-out auxiliary-condition items (out-of-domain/mixed) measured in LBS.}
    \label{fig:control_in_results}

\end{minipage}

\clearpage

\begin{minipage}[t]{1.0\textwidth}
    \centering
    
    \centering
    \includegraphics[width=\textwidth]{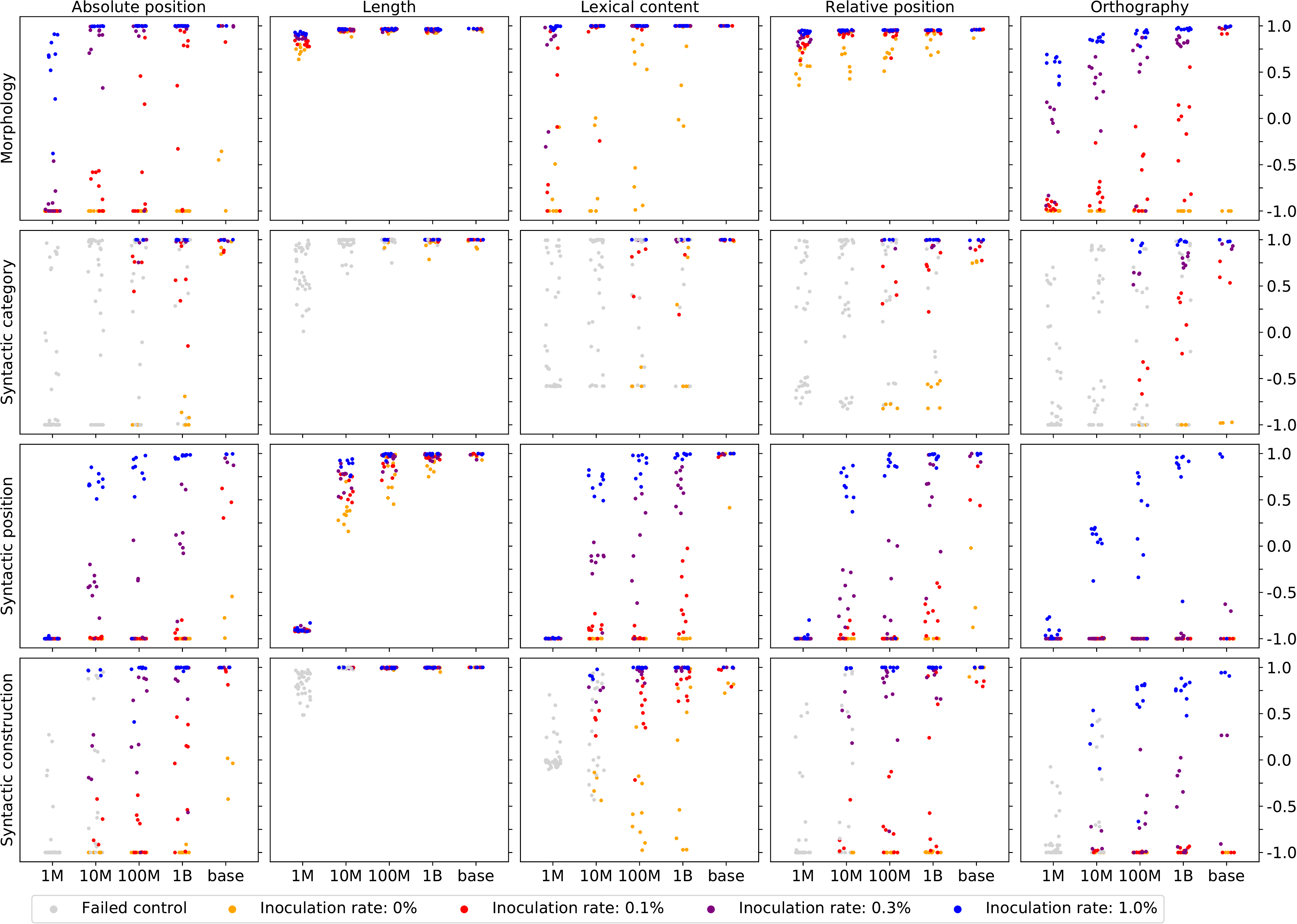}
    \captionof{figure}{Results on the held-out inoculation-condition items (in-domain/unmixed) measured in LBS.}
    \label{fig:control_out_results}
    
\end{minipage}

\clearpage

\section{Black and white versions of Fig. 4}

\begin{minipage}[t]{1.0\textwidth}
    \centering
    
    \includegraphics[width=\textwidth]{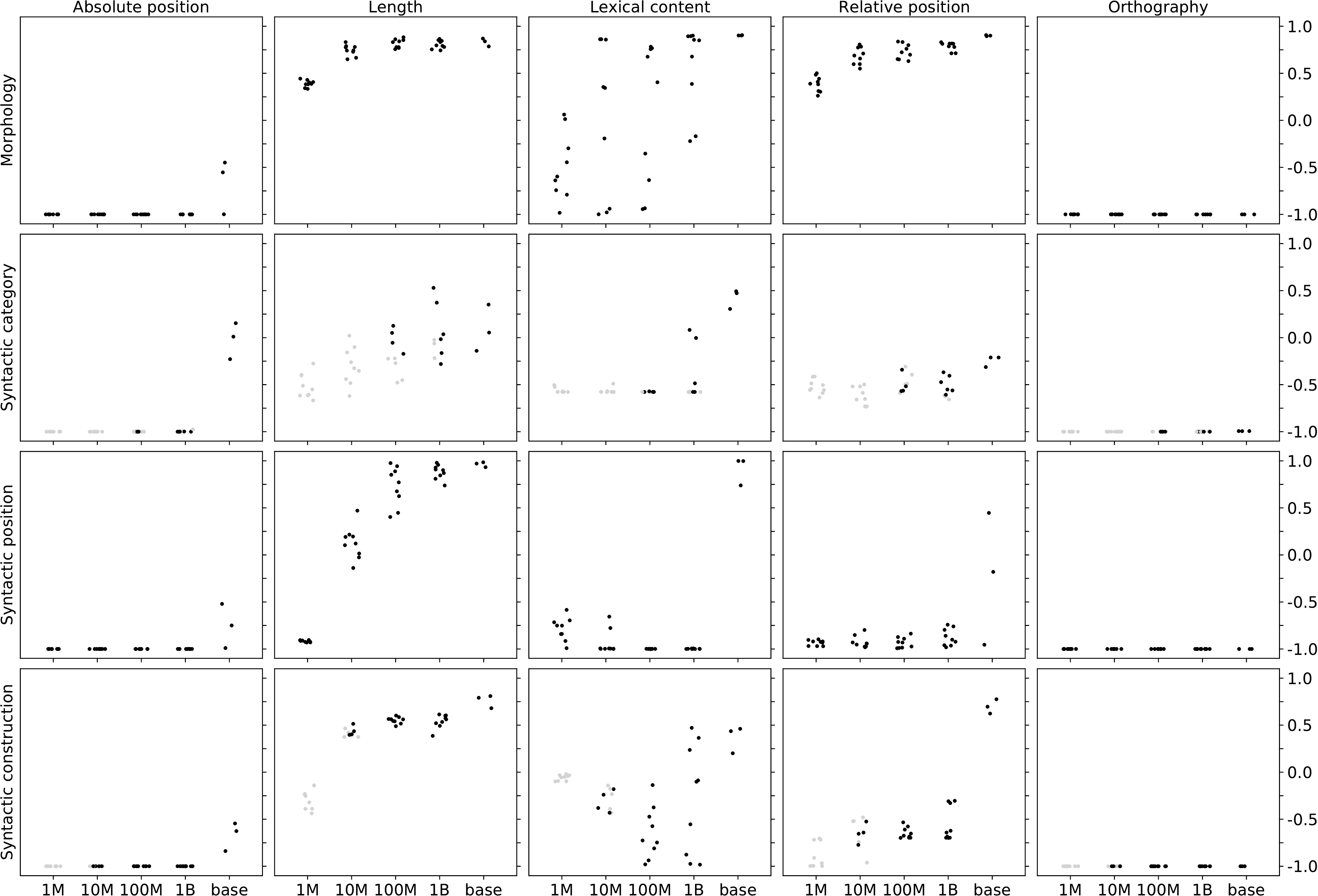}
    \captionof{figure}{Results of the mixed binary classification tasks with no inoculation data.}
    \label{fig:ino0_results}

    \centering
    \includegraphics[width=\textwidth]{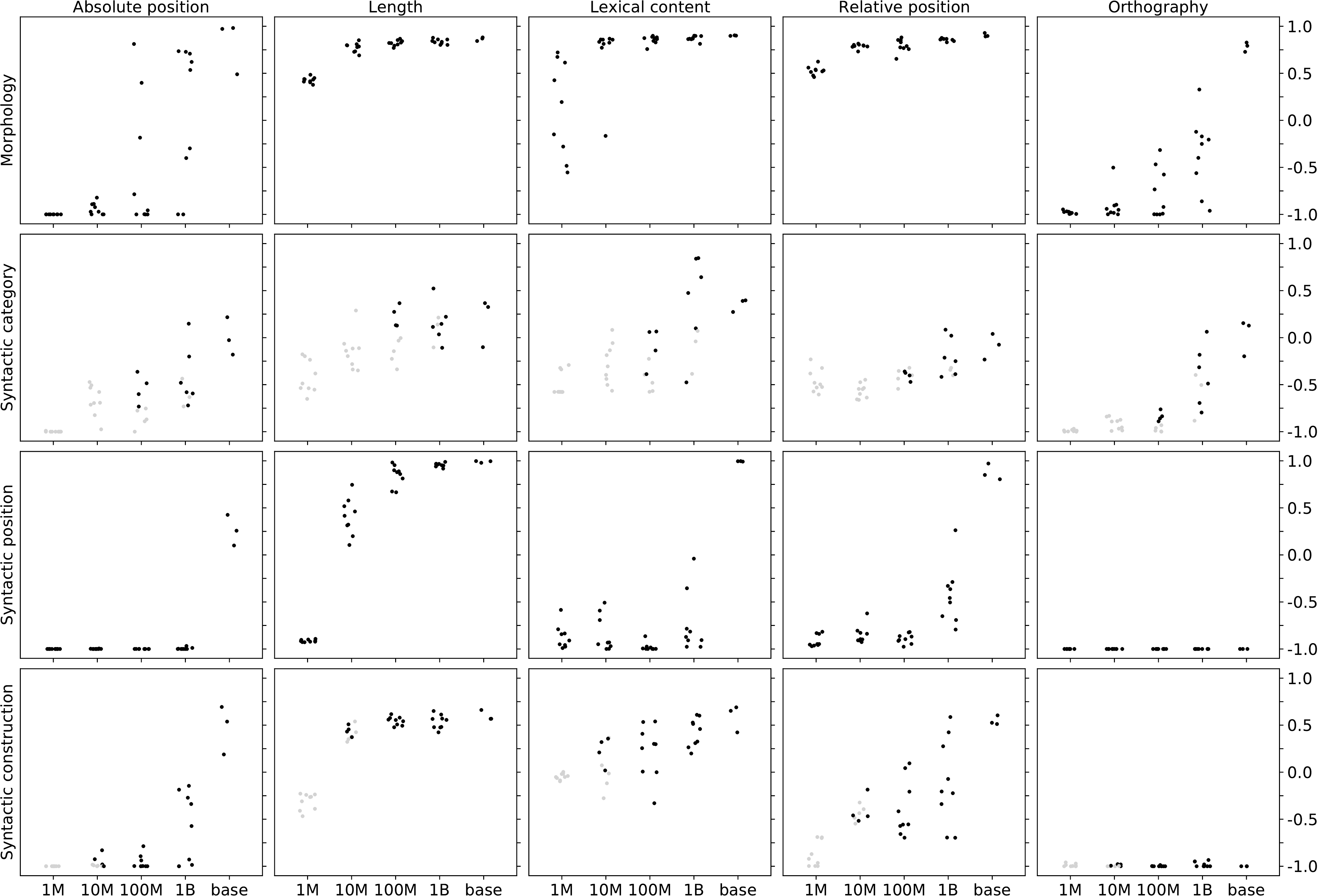}
    \captionof{figure}{Results of the mixed binary classification tasks with 0.1\% inoculation data.}
    \label{fig:ino001_results}

\end{minipage}

\clearpage

\begin{minipage}[t]{1.0\textwidth}
    \centering
    \includegraphics[width=\textwidth]{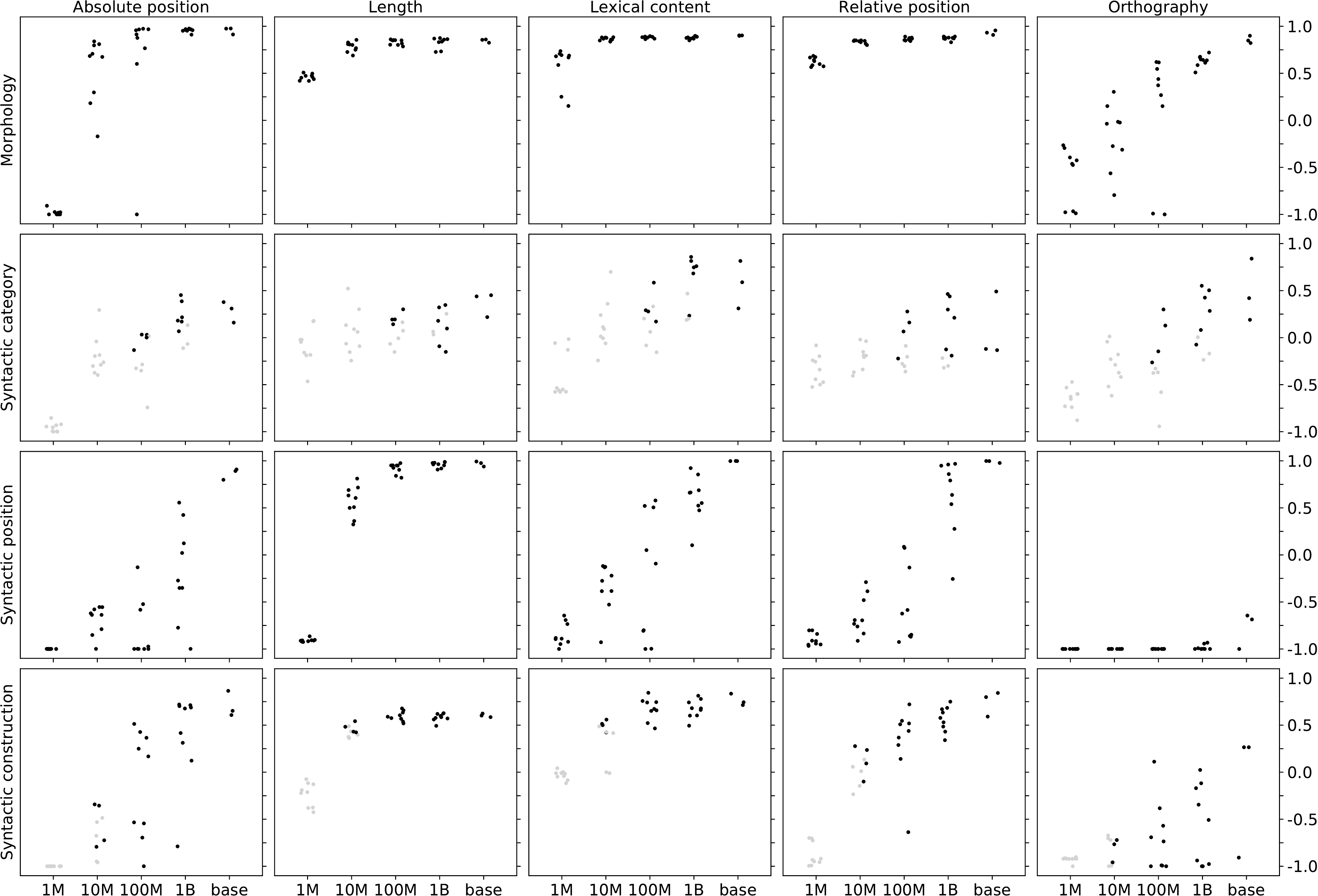}
    \captionof{figure}{Results of the mixed binary classification tasks with 0.3\% inoculation data.}
    \label{fig:ino003_results}

    \includegraphics[width=\textwidth]{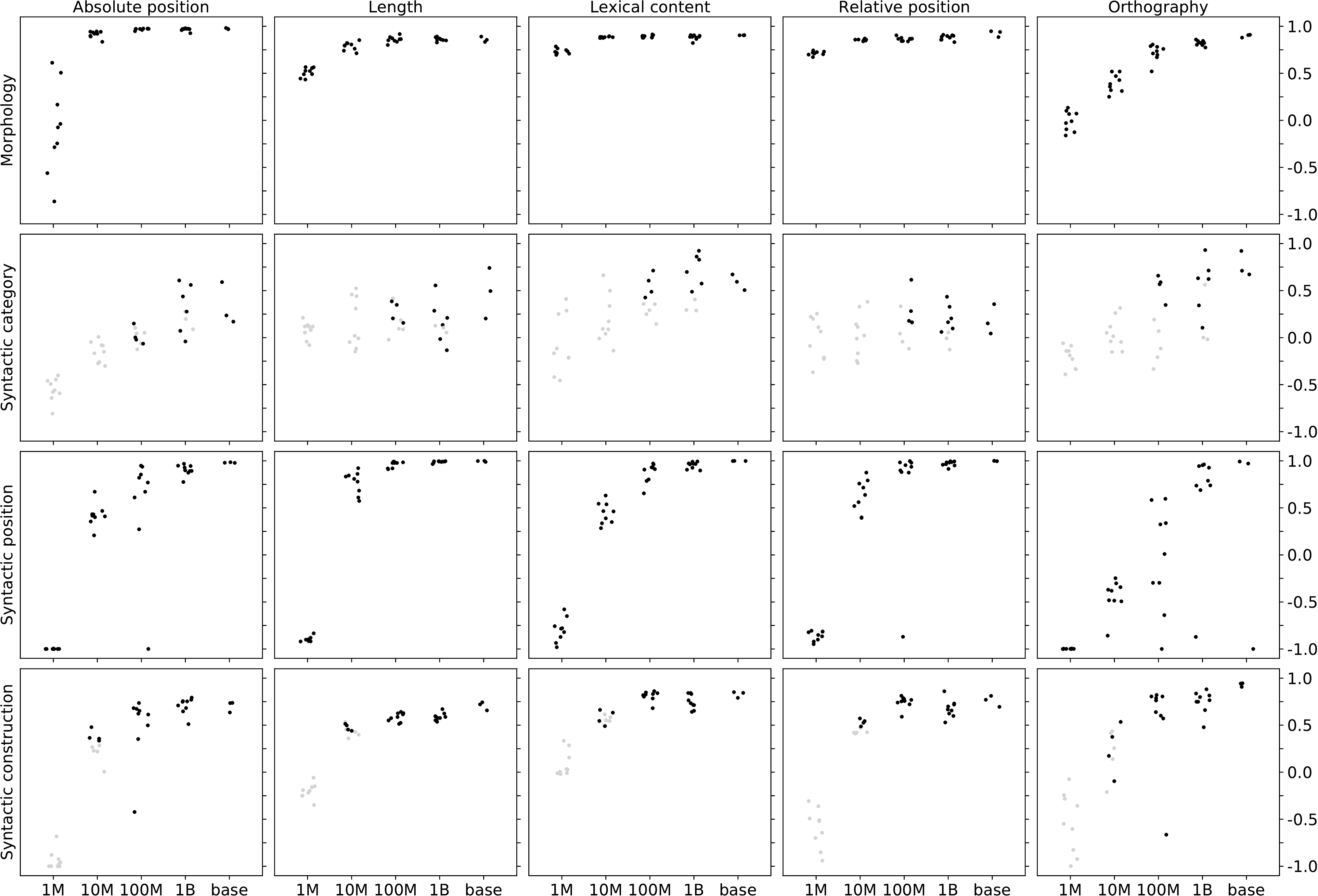}
    \captionof{figure}{Results of the mixed binary classification tasks with 1\% inoculation data.}
    \label{fig:ino01_results}
    
\end{minipage}

\end{document}